\documentclass[10pt,journal]{IEEEtran}
\usepackage{cite}
\usepackage{amsmath,amssymb,amsfonts}
\usepackage{graphicx}
\usepackage{booktabs}
\usepackage{multirow}
\usepackage[hidelinks]{hyperref}
\usepackage{orcidlink}
\usepackage[capitalise]{cleveref}
\crefname{figure}{Fig.}{Figs.}
\Crefname{figure}{Fig.}{Figs.}
\crefname{table}{Table}{Tables}
\Crefname{table}{Table}{Tables}
\usepackage{textcomp}

\DeclareMathOperator*{\Top}{Top}

\def\BibTeX{{\rm B\kern-.05em{\sc i\kern-.025em b}\kern-.08em
    T\kern-.1667em\lower.7ex\hbox{E}\kern-.125emX}}
\begin{document}

\title{Training-Free Out-of-Distribution Detection for Pathology Whole-Slide Images}

 \author{Sabri~Mustafa~Kahya, Richard~R.~Chen, Muhammet~Sami~Yavuz, Jerry~Jierui~Lou, Akanimoh~Adeleye, Hacı~Ali~Kahya, and Jana~Lipkova}

\maketitle

\begin{abstract}
Safe deployment of AI methods in medicine requires robust guardrails that detect when input data deviate from the training distribution to ensure that models provide predictions only within their scope of expertise and abstain otherwise. Out-of-distribution (OOD) detection can provide such safeguards and is extensively studied in general computer vision. Yet, it remains underdeveloped in computational pathology, where gigapixel whole-slide images (WSIs), subtle differences between disease subtypes, and variability in tissue preparation pose unique challenges for conventional OOD methods. We propose ZIO, a training-free, multimodal OOD detector for pathology WSIs that leverages vision--language pathology foundation models (FMs). ZIO constructs text and visual prototypes of in-distribution classes and integrates their complementary information through a prototype shrinkage mechanism to derive OOD scores. We provide the ZIO formulation for both slide- and patch-level FMs. We evaluate ZIO across diverse clinically relevant domain shifts, including rare diseases and near-OOD settings. Extensive evaluation of over 14,700 WSIs from five independent consortia shows that ZIO consistently outperforms both unimodal prototypes and 40 state-of-the-art OOD methods. These results demonstrate the benefits of multimodal representation for OOD detection and pave the way towards safer AI deployment in clinical practice. The code is available at \url{https://github.com/muskahya/ZIO}.
\end{abstract}


\section{Introduction}
\label{sec:introduction}

\IEEEPARstart{A}{I} models often operate under a closed-world assumption, expecting that the data seen at inference resemble the training data. In clinical settings, however, models may encounter rare or atypical cases. Without explicit guardrails, a model trained to predict a fixed set of diagnoses is forced to assign one of the predefined labels even when confronted with different anatomy or previously unseen conditions. Safe clinical deployment thus requires mechanisms that detect when input data deviates from the training distribution, enabling models to abstain rather than produce confident but incorrect predictions~\cite{rajpurkar2022ai,begoli2019need,zadorozhny2022ood,zimmerer2022mood,dolezal2022uncertainty}. This need is particularly acute in pathology, where subtle disease variations, differences in tissue preparation, staining protocols, scanner hardware, and patient demographics can induce distribution shifts that impact model performance~\cite{tellez2019quantifying,stacke2021measuring,chen2023algorithmic}.

Out-of-distribution (OOD) detection addresses this challenge by flagging inputs that deviate from the in-distribution (ID) training data. Unlike standard binary classification, OOD detection is typically developed using ID data only, since it is infeasible to anticipate all OOD types the model may encounter. Despite substantial progress in OOD detection for natural images~\cite{hendrycks2017baseline,liang2018enhancing,liu2020energy,hendrycks2022scaling}, the problem relatively remains unexplored in digital pathology. This gap stems from the unique complexity of gigapixel WSIs: a single slide may contain diverse tissue constituents (e.g., stroma, necrosis, inflammation) and even mixtures of ID and OOD-like regions (see \Cref{fig:pathology-details}). Consequently, most existing pathology OOD studies are patch-centric~\cite{linmans2024diffusion,oh2024we,pocevivciute2025out,linmans2020efficient,thagaard2020can}, relying on selected regions of interest that do not capture the heterogeneity of WSIs. While these studies demonstrate the feasibility of OOD detection, they are typically restricted to far-OOD settings (e.g. different organs), and often require expensive pretraining, which limits their application to WSIs and domain shifts presented in clinical practice.

Recently, zero-shot and few-shot OOD detection methods have emerged in computer vision as a new class of training-free approaches~\cite{esmaeilpour2022zero,wang2023clipn,ming2022delving,fu2025clipscope}. These methods rely on multimodal vision-text foundation models (FMs) trained with contrastive learning~\cite{radford2021learning} to build lightweight classifiers from text prompts (zero-shot) or a few exemplars (low-shot) in the embedding space~\cite{esmaeilpour2022zero,wang2023clipn,fu2025clipscope,miyai2023locoop,bai2024id}. While these methods have demonstrated strong performance on natural images, their application to computational pathology remains largely unexplored. At the same time, several multimodal vision-text FMs have been developed for computational pathology~\cite{lu2024visual,ding2025multimodal,li2025multi,huang2023visual,neidlinger2025benchmarking} over the last three years. These developments open new opportunities to translate the benefits of training-free OOD detection into computational pathology.

Here, we introduce ZIO (Zero-Shot-Informed One-Shot), a multimodal, training-free OOD detection framework that operates on entire WSIs and can be applied without backbone fine-tuning, calibration, or additional OOD data. ZIO constructs visual and text prototypes for each ID class and integrates their complementary information through prototype shrinkage to enable multimodal low-shot OOD detection in WSIs (\cref{fig:high-level-inference}). To be compatible with different types of FMs used in pathology, we provide the ZIO formulation for both slide- and patch-level FMs. Moreover, for patch-level FMs, we provide a new mechanism to determine which image regions correspond to the given text prompts in the WSI prior to multimodal fusion. We demonstrate that unimodal image and text prototypes provide strong baselines for OOD detection, but struggle under different types of domain shifts. By integrating complementary information from both modalities, ZIO overcomes these limitations and achieves robust performance across diverse domain shifts. We performed extensive evaluation across multiple clinically relevant domain shifts, including near-OOD and rare diseases with limited data, where many standard OOD methods struggle. Evaluation of over 14,700 WSIs from five independent consortia shows that ZIO outperforms the unimodal baselines and 40 state-of-the-art OOD detection methods. To our knowledge, this is the first work to demonstrate the feasibility of training-free multimodal OOD detection in pathology WSIs and underscores the benefits of multimodal representation for this task. This work paves the way towards safer deployment of AI in clinical practice.



\section{Related Work}
\label{sec:related_work}

\begin{figure}[t]
\centerline{\includegraphics[width=0.95\columnwidth]{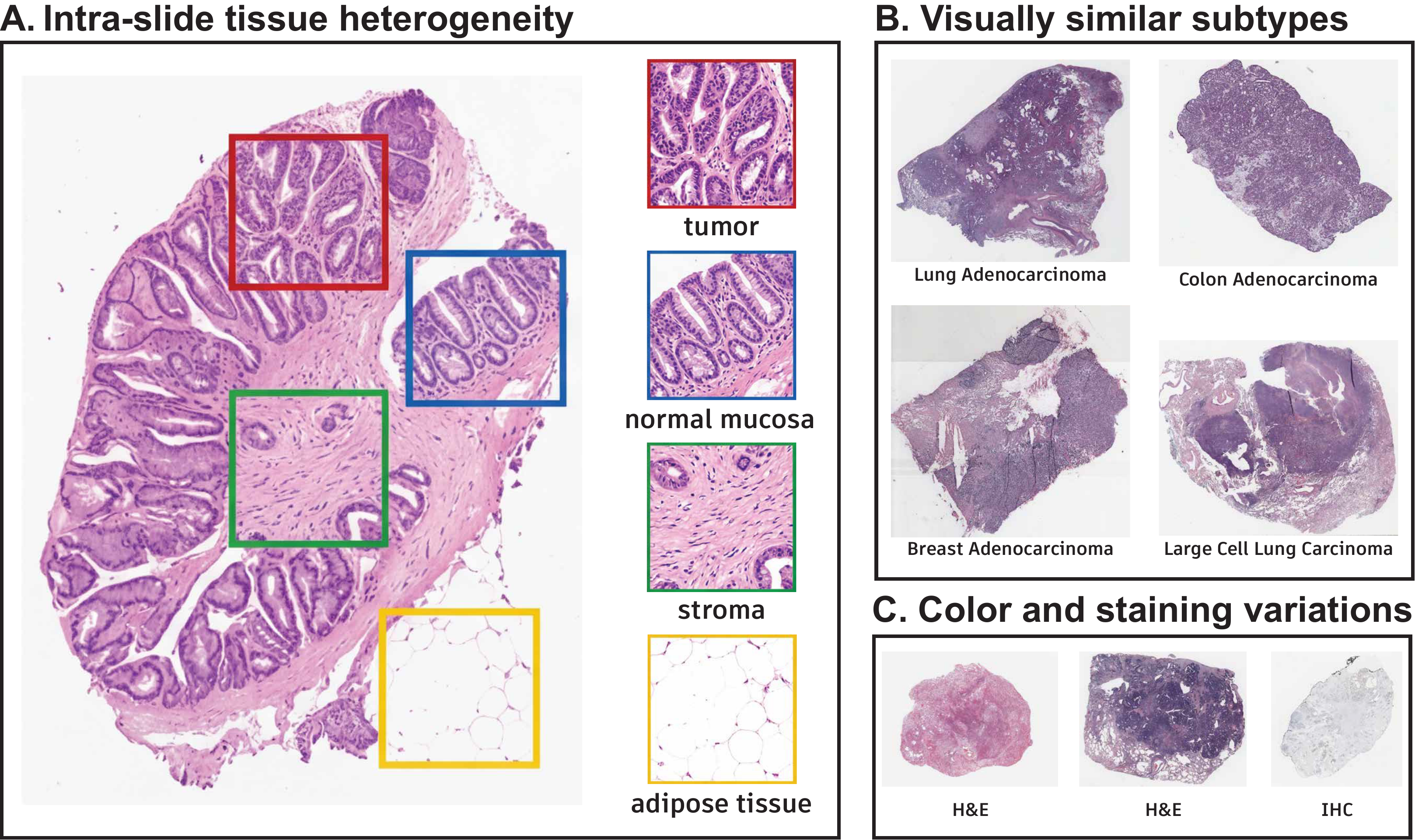}}
\caption{\textbf{Challenges in WSI-level OOD detection in pathology.}
(A) A single WSI can exhibit substantial intra-slide heterogeneity; example regions include tumor, normal mucosa, stroma, and adipose tissue.
(B) Histologically similar tumor types share highly similar morphology across different organs.
(C) Color and staining variability across slides (H\&E across scanners/centers and IHC) can induce strong covariate shift even for the same diagnostic categories.}
\label{fig:pathology-details}
\vspace{-3mm}
\end{figure}

\noindent\textbf{Computational pathology and foundation models.} Computational pathology develops algorithms that learn from digitized histopathology WSIs to support diagnosis, prognosis, and treatment selection~\cite{campanella2019clinical}. WSIs are gigapixel-resolution scans with substantial tissue heterogeneity within a slide, as shown in \cref{fig:pathology-details}. In practice, a WSI is usually partitioned into patches, encoded by a feature extractor, and aggregated into a slide-level representation using Multiple Instance Learning (MIL) methods. Attention-based MIL (ABMIL)~\cite{ilse2018attention,lu2021data} has become a standard aggregation strategy for downstream WSI tasks. FMs have recently advanced representation learning at both patch and slide levels~\cite{lipkova2024age}. Patch-level encoders pretrained with self-supervised learning on large histopathology corpora provide general purpose embeddings for image patches that can be used as a backbone for diverse downstream tasks~\cite{vorontsov2024foundation,chen2024towards}. Slide-level FMs capture global context by processing patches jointly or hierarchically, to directly output slide-level embeddings~\cite{wang2024pathology,xu2024gigapath}. Multimodal vision--language pathology FMs use CLIP-style contrastive learning~\cite{radford2021learning} to align histology images with text reports in the feature space, enabling zero-shot predictions and cross-modal retrieval. Representative examples of vision-language pathology FMs include PLIP~\cite{huang2023visual}, QUILT-Net~\cite{ikezogwo2023quilt}, MUSK~\cite{xiang2025vision}, PRISM~\cite{shaikovski2024prism}, CONCH~\cite{lu2024visual}, or TITAN~\cite{ding2025multimodal}. For more details see the recent review articles \cite{li2026survey,li2025multi}.

\noindent\textbf{OOD detection.} The objective of OOD detection is to distinguish ID and OOD samples~\cite{hood,dong2024multiood,mcrood,lidar}, as illustrated in \cref{fig:high-level-inference}A. Unlike a typical supervised binary classification task, OOD detectors are often developed using only the ID samples, since it is infeasible to anticipate all possible OOD cases. Existing OOD detection methods are typically applied on top of a trained ID model, most often using standard backbones such as ResNet-18 and ResNet-34. Below we provide an overview of different types of OOD methods, while more details are given in the review articles \cite{yang2024generalized,yang2022openood}.

\noindent\textbf{Post-hoc methods} enable OOD detection without retraining of the ID model. Early approaches leveraged confidence-based scores, where MSP~\cite{hendrycks2017baseline} established the baseline by using maximum softmax probabilities. ODIN~\cite{liang2018enhancing} enhances this through temperature scaling and input perturbations. Energy-based detection~\cite{liu2020energy} applies the LogSumExp function to logits, demonstrating that in-distribution (ID) samples exhibit lower energy scores. MaxLogit~\cite{hendrycks2022scaling} directly operates on logits and is based on the observation that ID samples tend to have higher maximum logit values than OODs.
\\
\noindent\textbf{Training-based methods} incorporate OOD awareness during model training. This includes Outlier Exposure (OE) approach \cite{b8}, which uses auxiliary outlier datasets and jointly trains the model on both ID and subset of OOD samples for better detection. Methods without OE modify the training objective or model architecture. For instance, CIDER~\cite{ming2023cider} optimizes hyperspherical embeddings through dispersion and compactness losses, increasing angular separation between classes.

\noindent\textbf{Zero-shot and few-shot OOD detection.}
Vision--language FMs trained with contrastive learning (e.g., CLIP~\cite{radford2021learning}) have enabled zero-shot OOD detection by building text-prompt prototypes for ID classes and deriving OOD scores from image--text similarity in the joint embedding space~\cite{esmaeilpour2022zero,ming2022delving,wang2023clipn,fu2025clipscope}. Few-shot variants further leverage limited visual exemplars or prompt adaptation to improve separability~\cite{miyai2023locoop,bai2024id}. However, most prior works were designed for natural images. 



\noindent\textbf{OOD detection in pathology} remains relatively unexplored in comparison to studies on natural images. Most early methods \cite{linmans2024diffusion,oh2024we,pocevivciute2025out,linmans2020efficient,thagaard2020can} operate on small regions of interest or image patches. However, this approach is not suitable for clinical deployment, since a single WSI can contain both ID and OOD samples.
Linmans et al. \cite{linmans2023predictive} proposed a way to aggregate information from the image patches to enable a slide-level OOD detection.
However, this approach is based on thresholding of the entropy scores computed across patches, where the choice of the threshold is task-dependent, making the model less flexible for broader deployment. Recently, StaDis \cite{zhang2025stadis} and SCULPT \cite{sun2026sculpt} enabled slide-level OOD detection within MIL frameworks using perturbation-based stability scoring and causal prompt tuning, respectively. The early works \cite{linmans2023predictive,linmans2020efficient,pocevivciute2025out} usually focused on relatively simple settings; the models typically consider only a single ID class (e.g. presence/absence of tumor) and are tested for far-OOD cases with very distinct tissue morphology, e.g. prostate cancer as ID and healthy colon tissue as OOD \cite{linmans2023predictive,linmans2020efficient,pocevivciute2025out}. The recent studies \cite{sun2026sculpt,zhang2025stadis} further included more clinically challenge domain shift, such as different cancer types within the same primary tissue. However these models are tested only on a handful of cancer types, and thus their generalization across different diseases remain underexplored. Moreover Pocevi{\v{c}}i{\=u}t{\.e} \cite{pocevivciute2025out}, showed that existing OOD methods in pathology do not generalize well to data from external cohorts. These prior works show the feasibility but also underscore the challenges of OOD detection in histology WSIs.

\section{Method}
\label{sec:method}

\begin{figure}[t]
\centerline{\includegraphics[width=1.0\columnwidth]{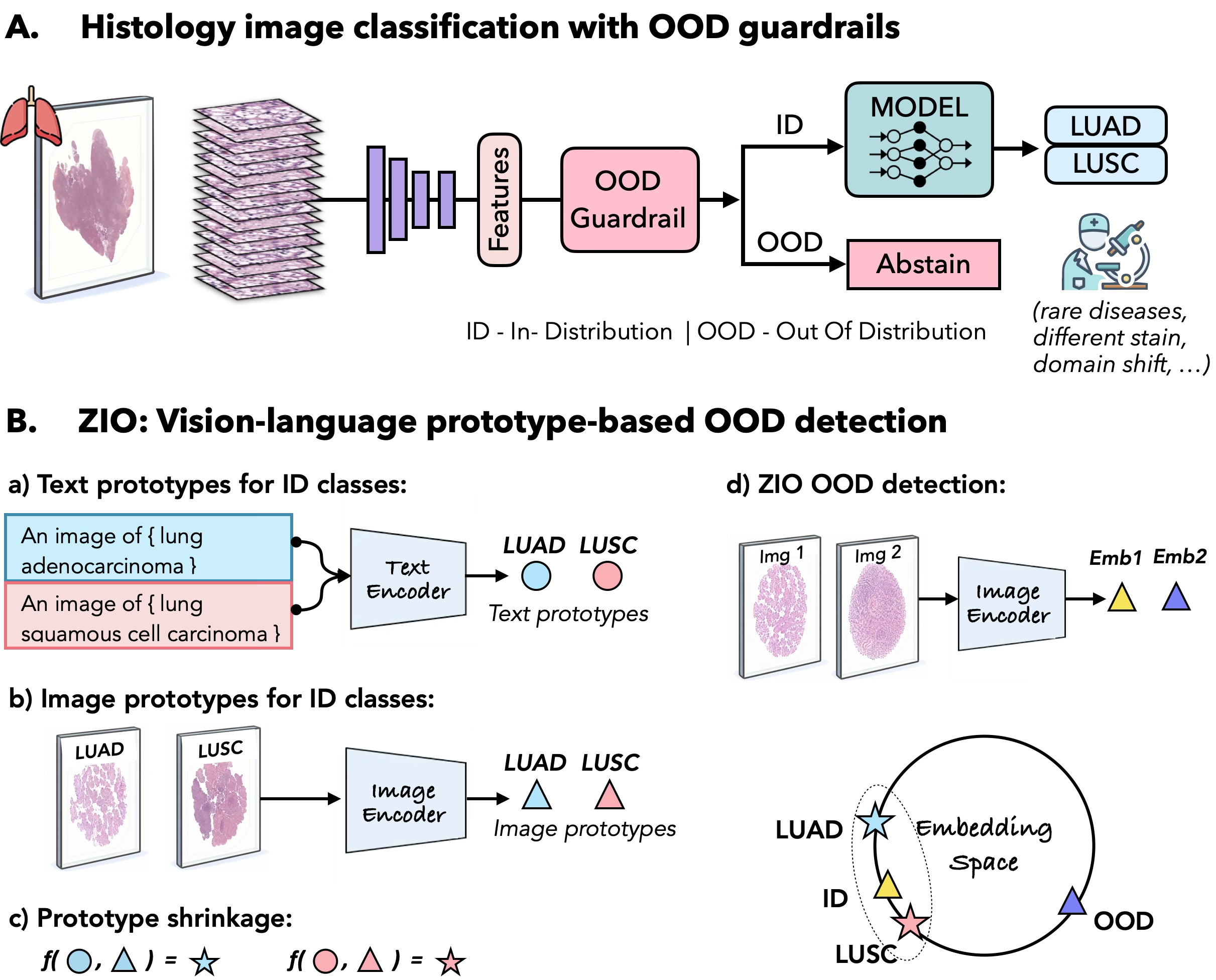}}
\caption{\textbf{High-level study overview.} 
(A) Example of AI-based classification of lung cancer into LUAD and LUSC subtypes, while using OOD safeguards. OOD detection flags cases that deviate from model's expertise, such as rare disease or different domain shifts, to ensure safe model deployment. (B) High-level overview of the ZIO OOD detection using multimodal image and text prototypes.}
\label{fig:high-level-inference}
\vspace{-3mm}
\end{figure}

\noindent\textbf{Problem setup.}
OOD detection is formulated as a binary classification problem where the objective is to distinguish ID samples from OOD ones (see \cref{fig:high-level-inference}A). Unlike standard supervised classification, the OOD detector has only access to ID data during model development. The goal is to learn a scoring function that captures the characteristics of the ID samples and flags data that deviates from this distribution. A typical OOD detector first assigns a score to each input sample $x$ using a scoring function $s$, and then classifies the sample based on a predefined threshold $\tau$: if $s(x) < \tau$, the sample is classified as OOD; otherwise ID.

\vspace{2mm}
\noindent\textbf{Notation.} Let $\mathcal{Y}=\{1,\dots,C\}$ denote the ID classes. For each WSI $x$, we extract representations using a FM. The slide-level FM produces a single embedding $h\in\mathbb{R}^{D_{sl}}$, while patch-level FM yields a set of patch embeddings $U=\{u_p\in\mathbb{R}^{D_{pl}}\}_{p=1}^{P}$, where $P$ denotes the number of patches in a given WSI $x$. The proposed ZIO method is compatible with any vision-language FM. Here we use TITAN~\cite{ding2025multimodal}, as a slide-level FM and CONCH~\cite{lu2024visual} as a patch-level FM. Under this setup, the embedding dimensions are $D:= D_{sl} = 768$ and $D:= D_{pl} = 512$, respectively. We denote the text-only prototypes as \textbf{ZS}, due to their conceptual link to zero-shot learning, and the image-only prototypes as \textbf{OS}, as they are derived from one-shot visual exemplars. \textbf{ZIO} and its unimodal variants, ZS and OS, each construct a prototype matrix
$W = [w_1, \dots, w_C] \in \mathbb{R}^{D \times C}$,
where $w_c$ denotes the class-$c$ prototype. Let $\mathrm{norm}(\cdot)$ denote $\ell_2$ normalization.

\vspace{-2mm}
\subsection{ZIO: Multimodal OOD Detection in WSIs}
\label{sec:zio}
\begin{figure*}[t!]
\centerline{\includegraphics[width=0.75\linewidth]{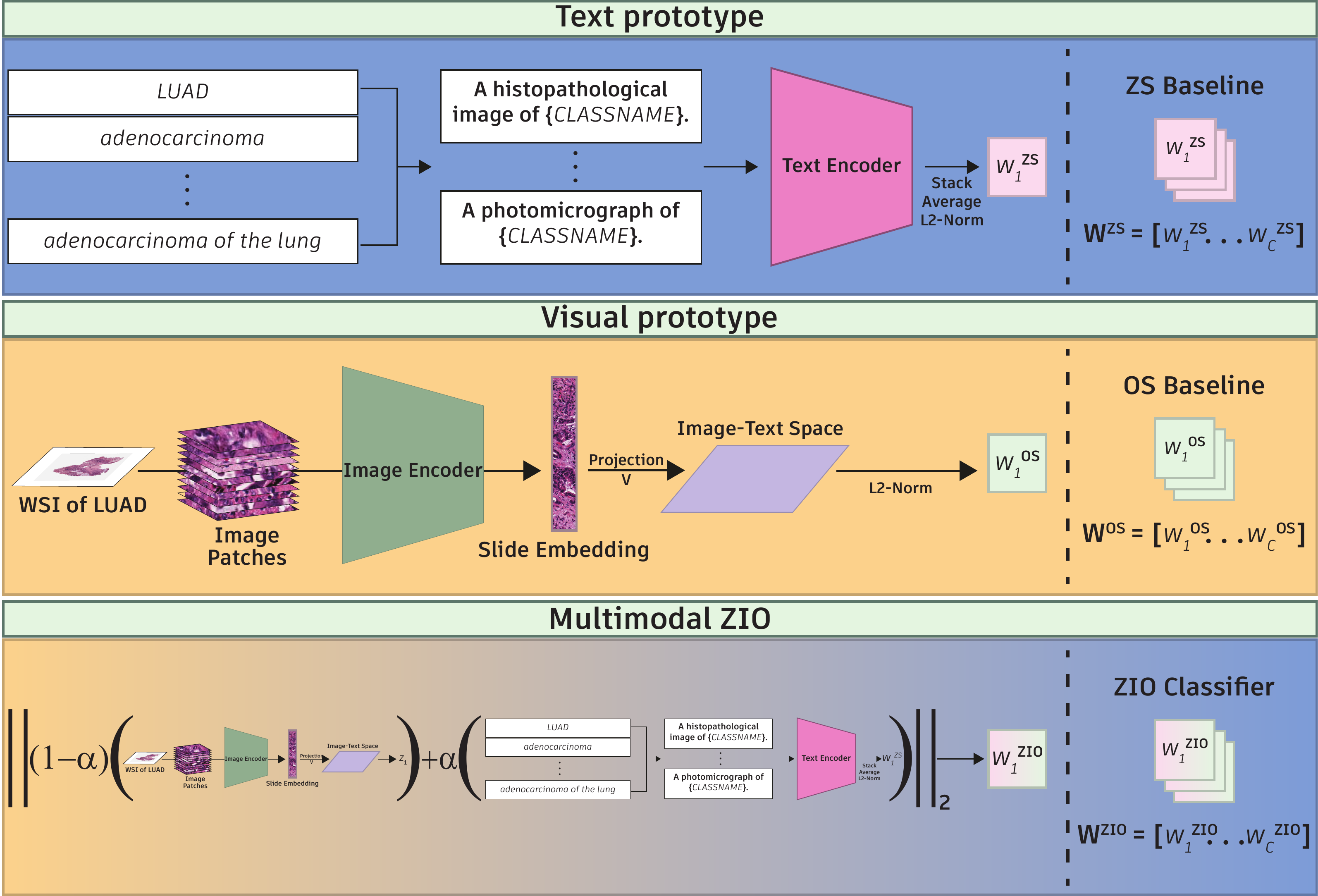}}
\caption{\textbf{ZIO prototype construction and fusion.} Top: text prototype used as the unimodal ZS baseline, built from class names/synonyms and prompt templates via the FM text encoder. Middle: visual prototype used as the unimodal OS baseline, built from a single WSI per class via the FM image encoder. Bottom: ZIO fuses the text and visual prototypes via prototype shrinkage to form a multimodal image-text prototype.}
\label{fig:model}
\end{figure*}

\noindent ZIO constructs text and visual prototypes and integrates them via prototype shrinkage to provide multimodal OOD detection. We first describe the construction of the individual prototypes, followed by the fusion mechanism and the resulting scoring function. A high-level overview of the ZIO model is provided in \cref{fig:high-level-inference}B, with model specifics shown in \cref{fig:model}.

\vspace{2mm}
\noindent\textbf{ZS: text prototypes.}
The text prototypes use only the text description of the ID classes, no images. For each ID class $c$, given a set of class name synonyms $\mathcal{S}_c$ (e.g., \textit{lung adenocarcinoma, LUAD}) and prompt templates $\mathcal{T}$ (e.g., ``\textit{the pathology image shows \texttt{CLASSNAME}}''), we build the prototype matrix $W^{\text{ZS}}=[w_1^{\text{ZS}},\dots,w_C^{\text{ZS}}]$ using the FM's text encoder $f_{\text{text}}$ as:
\[
\begin{aligned}
    w_c^{\text{ZS}} \;=\; \mathrm{norm}\!\left(\frac{1}{|\mathcal{S}_c||\mathcal{T}|}\sum_{s\in\mathcal{S}_c}\sum_{t\in\mathcal{T}} \mathrm{norm}\!\big(f_{\text{text}}(t[s])\big)\right).
\end{aligned}
\]

\noindent\textbf{OS: visual prototypes.}
Using a single WSI for each ID class, we construct image prototypes
$W^{\text{OS}} = [w_1^{\text{OS}}, \dots, w_C^{\text{OS}}]$.
We describe the procedure separately for slide-level and patch-level visual representations.

\noindent\textit{i) Slide-level.} Given a WSI of class $c$, the slide-level FM provides a single slide-level embedding $h_c\in\mathbb{R}^{D}$. Using the pretrained projection head $V\in\mathbb{R}^{D\times D}$ of the FM, we map the raw image features $h_c$ into the joint image--text embedding space established during contrastive pretraining, i.e., $z_c=\mathrm{norm}(h_c^\top V)\in\mathbb{R}^{D}$. The visual prototype is then defined as
\vspace{-2mm}
\[
w_c^{\text{OS}}=\mathrm{norm}(z_c).
\]

\noindent\textit{ii) Patch-level.} Patch-level FM produces embeddings $U^{(c)}=\{u_p^{(c)} \}_{p=1}^{P}$ for each image patch of a given WSI of class $c$. Using the projection head $A\in\mathbb{R}^{D\times D}$ of the FM, we map each patch embedding into the joint image-text feature space as follows: $z^{(c)}_p=\mathrm{norm}(u_p^{(c)\top} A)\in\mathbb{R}^{D}$. The visual prototype of a given WSI is obtained by mean pooling over all patches, followed by $\ell_2$ normalization:
\[
\bar z^{(c)} = \frac{1}{P}\sum_{p=1}^{P} z_p^{(c)};\;\text{ }
w_c^{\text{OS}}=\mathrm{norm}(\bar z^{(c)}).
\]

\noindent\textbf{ZIO: multimodal fusion.}
ZIO performs prototype shrinkage to build the multimodal visual-text prototypes $W^{\text{ZIO}}=[w_1^{\text{ZIO}},\dots,w_C^{\text{ZIO}}]$ as follows:

\noindent\textit{i) Slide-level.} In the slide-level formulation, given the text prototype $w_c^{\text{ZS}}$ and the one-shot slide-level prototype $w_c^{\text{OS}}$, we can directly compute the prototype shrinkage as follows:
\[
w_c^{\text{ZIO}}=\mathrm{norm}\!\big((1-\alpha)\,w_c^{\text{OS}} + \alpha\,w_c^{\text{ZS}}\big),
\quad \alpha\in[0,1].
\]

\noindent\textit{ii) Patch-level.} In pathology, the diagnostically relevant tissue often occupies only a small fraction of a WSI, while the remaining regions include other tissue constituents (see \cref{fig:pathology-details}). In the patch-level model formulation, the text-based prototype $w_c^{\text{ZS}}$ corresponds to diagnostic class $c$. However, it is not clear which image patches in the WSI are representative of a given class $c$. Consequently, it is non-trivial to determine how visual and text prototypes should be aligned for the multimodal fusion.
To address this challenge, we introduce a ZS-guided top-$j$ filtering mechanism that allows us to select the most informative image patches and discard the signal from non-informative patches prior to multimodal fusion. This is achieved through the following two stages.

\textit{Stage 1: ZS-guided patch selection.} Given the projected image patch embeddings
$\{z_p^{(c)}\}_{p=1}^P$ for a WSI of class $c$, we compute their cosine similarity with the corresponding text prototype $w_c^{\text{ZS}}$. The similarity score for patch $p$ is defined as
$\text{score}_p^{(c)} = z_p^{(c)\top} w_c^{\text{ZS}}$.
We then select the set $\mathcal{I}_j$ corresponding to the top-$j$ patches with the highest similarity scores:
\[
\mathcal{I}_j = \operatorname{Top}\text{-}j\big(\{\text{score}_p^{(c)}\}_{p=1}^P\big).
\]

\textit{Stage 2: Multimodal fusion.} We define the visual prototype of class $c$ by computing the average of the top-$j$ informative patches, followed by normalization.
\[
\tilde z^{(c)} = \frac{1}{j}\sum_{p\in\mathcal{I}_j} z_p^{(c)},\quad
w_c^{\text{pre}}=\mathrm{norm}(\tilde z^{(c)}).
\]
The multimodal prototype is then computed through prototype shrinkage as:
\[
w_c^{\text{ZIO}}=\mathrm{norm}\!\big((1-\alpha)\,w_c^{\text{pre}} + \alpha\,w_c^{\text{ZS}}\big),
\quad \alpha\in[0,1].
\]

\noindent\textbf{OOD detection.}
\label{sec:pooling}
Given an input WSI $x$, the OOD score $s(x)$ is computed using the prototype matrix $W$, where $W^{\text{ZS}}$, $W^{\text{OS}}$, and $W^{\text{ZIO}}$ correspond to the ZS, OS baselines, and multimodal ZIO detector, respectively.

\noindent\textit{i) Slide-level.}
We compute the logits for OOD detection via cosine similarity between the slide embedding $z$ and the prototype matrix $W$, as follows: $\ell=z^\top W$. At this point, any standard logit-based OOD scoring function can be applied. In this work, we use \textit{MaxLogit}, which defines the scoring function as $s_{\max}(x) = \max_{c} \ell_c$. We intentionally adopt vanilla MaxLogit for its simplicity, allowing us to isolate the effect of the proposed prototype-based formulation. Other logit-based OOD scoring methods (e.g., Energy, MSP (after softmax), or LogitNorm) could be used in the same manner.

\noindent\textit{ii) Patch-level.} For the input WSI, we compute per-patch logits $\ell_{p}=z_p^\top W$. We use top-$j$ pooling to select the patches with the highest per-class confidence:
\vspace{-1mm}
\[
\ell_c^{\text{pool}} \;=\; \frac{1}{j}\sum_{p\in \Top\text{-}j\{\ell_{p,c}\}_{p=1}^P} \ell_{p,c},
\]
MaxLogit is then used to compute the scoring function as $s_{\max}(x)=\max_c \ell_c^{\text{pool}}$.

\vspace{2mm}
\noindent\textbf{OOD decision rule and evaluation metrics:} Following the standard practice in OOD detection \cite{yang2024generalized,yang2022openood}, the threshold $\tau$ is determined by ensuring that $95\%$ of ID samples are correctly detected. We report common OOD detection metrics that compute performance across all values of $\tau$: Area Under the Receiver Operating Characteristic curve (AUROC), Area Under the Precision-Recall Curve when IDs are positives (AUPR\textsubscript{IN}), and when OODs are positives (AUPR\textsubscript{OUT}); and False Positive Rate at 95\% True Positive Rate (FPR95).

\noindent\textbf{Hyperparameters.}
For ZIO shrinkage, we use $\alpha=0.5$, assigning equal weight to the text and visual prototypes. For patch-level models, we set $j=10$ in all experiments. All feature vectors and prototype columns are $\ell_2$-normalized after projection. For methods that use visual exemplars (OS and ZIO), results are averaged over 30 randomly sampled one-shot episodes to assess robustness.

\noindent\textbf{Complexity.}
Per-slide complexity is $O(PC)$ for patch-FM (with $P$ patches and $C$ classes) and $O(C)$ for slide-FM.

\section{Experiments}
\label{sec:experiments}

\noindent\textbf{Dataset and preprocessing.}
We evaluate all OOD detectors on over 14,700 WSIs sourced from five public consortia: CPTAC~\cite{edwards2015cptac}, PLCO~\cite{gohagan2000prostate}, TCGA~\cite{weinstein2013cancer}, EBRAINS~\cite{amunts2016human}, and GTEx~\cite{gtex2020atlas}, as shown in \Cref{tab:ood_datasets}. We partition the data into different ID and OOD configurations to simulate scenarios with varying levels of difficulty, ranging from \textit{far-OOD} to \textit{very near-OOD}. Each configuration is described in the corresponding experiments section. All WSIs are processed using the TRIDENT~\cite{zhang2025standardizing} framework. The tissue regions are segmented and tiled into non-overlapping $512{\times}512$ patches at $20{\times}$ magnification. 

\vspace{1mm}
\noindent\textbf{Implementation of baseline models.} To provide a rigorous evaluation, ZIO is compared against 40 SOTA OOD detectors. All OOD methods operate on the exactly the same patch- and slide-level embeddings from the FMs as ZIO. This allows us to determine the benefits of ZIO against the baseline models, regardless of the choice of FM.
For the SOTA baselines, we follow the standard pipeline: train an ID classifier on top of these embeddings, then apply the corresponding OOD methods. For the slide-level formulation, we train a ResNet-18 classifier on top of slide-level embeddings. For the patch-level formulation, we first apply ABMIL~\cite{ilse2018attention} to aggregate patch features into a slide-level representation for a given downstream task, followed by a ResNet-18 classifier. Although in computational pathology it is common to use linear probing (e.g., logistic regression) or MLP layers after the slide-level embeddings (either from slide-level FM or after MIL), we opt for ResNet-18, because most of SOTA OOD detectors are designed and optimized for ResNet-style architectures. Moreover, several of these methods require access to intermediate features, penultimate-layer representations, gradients, or activations, which cannot be well established with linear probing. Since the resulting inputs to ResNet-18 are 1D embedding vectors, we reshape each embedding to a $1\times1\times D$ tensor and modify the first convolution to accept a single-channel input. After model training, we apply each OOD detector following its original implementation. For methods that require auxiliary OOD samples, such as OE methods, we use 516 WSIs from TCGA KIRC cohort, which are excluded from all reported OOD experiments.

{\setlength{\textfloatsep}{6pt plus 2pt minus 2pt}
\begin{table}[h]
\centering
\scriptsize 
\caption{\scriptsize\textbf{Overview of cohorts, datasets and WSI counts.}ALL LISTED DATASETS CONTAIN H\&E-STAINED WSIS EXCEPT PLCO-IHC, WHICH CONTAINS IHC-STAINED WSIS.}
\label{tab:ood_datasets}
\setlength{\tabcolsep}{4pt} 
\resizebox{\columnwidth}{!}{%
\begin{tabular}{@{}l r @{\hskip 12pt} l r @{\hskip 12pt} l r @{\hskip 12pt} l r @{\hskip 12pt} l r@{}}
\toprule
\multicolumn{2}{c}{CPTAC~\cite{edwards2015cptac}} &
\multicolumn{2}{c}{TCGA~\cite{weinstein2013cancer}}  &
\multicolumn{2}{c}{PLCO~\cite{gohagan2000prostate}}  &
\multicolumn{2}{c}{EBRAINS~\cite{amunts2016human}} &
\multicolumn{2}{c}{GTEx~\cite{gtex2020atlas}} \\
\cmidrule(r{6pt}){1-2} 
\cmidrule(lr{6pt}){3-4} 
\cmidrule(lr{6pt}){5-6} 
\cmidrule(lr{6pt}){7-8}
\cmidrule(l{6pt}){9-10}
LUAD & 398 & LUAD & 539 & LUAD & 399 & RB$_0$ & 652 & GTEx-near1 & 382 \\
LUSC & 405 & LUSC & 512 & LUSC & 224 & RB$_1$ & 142 & GTEx-near2 & 341 \\
BRCA & 650 & BRCA & 859 & BRCA & 860 & RB$_2$ & 507 & & \\
COAD & 370 & COAD & 458 & COAD & 846 & RB$_3$ & 2132 & & \\
GBM  & 525 & GBM  & 859 & IHC  & 434 &        &      & & \\
PDA  & 556 & PAAD & 209 & RARELUNG & 115      &        &      & & \\
UCEC & 882 & SKCM & 475 &      &      &        &      & & \\
\bottomrule
\end{tabular}}
\end{table}

\vspace{-2mm}
\subsection{Common Cancers: Very-far, Far, and Near-OOD}\label{sec:exp1}
The objective of this experiment is threefold:
(i) to evaluate the proposed ZIO method against its unimodal counterparts, ZS and OS; (ii) to assess whether the benefits of multimodal prototype fusion generalize across external cohorts and different malignancies; and (iii) to benchmark ZIO against 40 SOTA OOD baselines. The ID task is subtyping of non-small cell lung carcinoma (NSCLC) into the two most common subtypes: lung adenocarcinoma (LUAD) and lung squamous cell carcinoma (LUSC). The ID classifier is trained on set of 1,293 Hematoxylin and Eosin (H\&E)-stained WSIs (also excluded from all reported experiments) from CPTAC LUAD/LUSC cohort. The OOD detection methods are tested against different types of domain shifts. We consider \textit{(i) far-OOD} cases, which include H\&E-stained WSIs from cancers that are biologically and histologically distinct from lung tumors, such as glioblastoma (GBM), skin cutaneous melanoma (SKCM), or uterine corpus endometrial carcinoma (UCEC). The \textit{(ii) near-OOD} cases include adenocarcinomas originating from other organs, such as breast (BRCA), colon (COAD) or pancreatic (PAAD/PDA) cancers, which share the glandular morphology and adenocarcinoma characteristics with the ID LUAD class. These are significantly more challenging cases because the histology samples are usually taken from the tumor regions and thus they contain only minimal or no "healthy" tissue, while the adenocarcinoma histologic representation is almost identical across all cases. Additionally, we consider \textit{(iii) very far-OOD}, where Immunohistochemistry (IHC)-stained WSIs represent the OOD cases, which are visually distinct from H\&E-stained ID WSIs (\cref{fig:pathology-details}C). 

\begin{table*}[t]
\centering
\scriptsize
\setlength{\tabcolsep}{3pt}
\caption{\footnotesize\textbf{Performance across very-far, far, and near OOD scenarios}. ID: CPTAC-LUAD/LUSC. All metrics show $\textit{ZS} | \textit{OS} | \textit{ZIO}$}
\label{tab:titan_conch_zs_os_zio_ood_grouped}
\resizebox{\textwidth}{!}{
\begin{tabular}{@{}l l c c c c@{}}
\toprule
\textbf{OOD Scenario} & \textbf{Cohort Dataset} & \textbf{AUROC} & \textbf{AUIN} & \textbf{AUOUT} & \textbf{FPR95} \\
\midrule
\multicolumn{6}{l}{\textbf{Slide-Level}} \\
\midrule
\textit{Very-far OOD} & PLCO IHC & $99.96|97.10_{\pm4.61}|\mathbf{99.96}_{\pm0.06}$ & $\mathbf{99.85}|98.57_{\pm2.10}|\mathbf{99.85}_{\pm0.03}$ & $99.69|91.97_{\pm11.55}|\mathbf{99.70}_{\pm0.09}$ & $\mathbf{0.00}|16.57_{\pm29.77}|0.11_{\pm0.31}$ \\
\cmidrule(lr){1-6}
\multirow{4}{*}{\textit{Far OOD}} & CPTAC GBM & $97.25|96.78_{\pm3.67}|\mathbf{98.89}_{\pm0.67}$ & $98.25|97.94_{\pm2.28}|\mathbf{99.19}_{\pm0.43}$ & $95.04|94.06_{\pm6.26}|\mathbf{98.01}_{\pm1.04}$ & $13.71|16.46_{\pm18.51}|\mathbf{4.93}_{\pm4.66}$ \\
 & CPTAC UCEC & $87.94|90.20_{\pm7.52}|\mathbf{93.65}_{\pm2.98}$ & $88.87|91.04_{\pm6.92}|\mathbf{94.22}_{\pm2.76}$ & $87.11|88.98_{\pm7.71}|\mathbf{93.03}_{\pm3.01}$ & $56.24|44.66_{\pm21.23}|\mathbf{37.72}_{\pm10.95}$ \\
 & TCGA GBM & $97.17|97.39_{\pm3.13}|\mathbf{99.08}_{\pm0.81}$ & $97.58|97.77_{\pm2.44}|\mathbf{99.10}_{\pm0.62}$ & $96.39|96.29_{\pm4.41}|\mathbf{98.77}_{\pm1.11}$ & $13.04|14.87_{\pm20.21}|\mathbf{4.07}_{\pm6.49}$ \\
 & TCGA SKCM & $89.80|91.19_{\pm6.99}|\mathbf{94.51}_{\pm2.58}$ & $94.25|95.21_{\pm3.60}|\mathbf{96.96}_{\pm1.36}$ & $78.99|81.78_{\pm13.11}|\mathbf{88.66}_{\pm5.25}$ & $55.58|45.83_{\pm28.87}|\mathbf{34.10}_{\pm15.76}$ \\
\cmidrule(lr){1-6}
\multirow{8}{*}{\textit{Near OOD}} & CPTAC BRCA & $91.05|91.39_{\pm8.34}|\mathbf{95.12}_{\pm3.16}$ & $92.56|93.77_{\pm5.65}|\mathbf{96.20}_{\pm2.41}$ & $89.51|87.62_{\pm10.63}|\mathbf{93.66}_{\pm3.80}$ & $36.92|39.26_{\pm22.28}|\mathbf{25.12}_{\pm12.44}$ \\
 & CPTAC COAD & $96.79|97.10_{\pm2.84}|\mathbf{98.65}_{\pm0.75}$ & $98.25|98.60_{\pm1.26}|\mathbf{99.24}_{\pm0.34}$ & $93.66|92.26_{\pm6.88}|\mathbf{96.78}_{\pm2.11}$ & $11.89|14.53_{\pm15.70}|\mathbf{6.39}_{\pm3.65}$ \\
 & CPTAC PDA & $91.04|94.00_{\pm7.24}|\mathbf{96.57}_{\pm2.23}$ & $93.88|96.03_{\pm4.33}|\mathbf{97.66}_{\pm1.48}$ & $85.91|90.42_{\pm11.13}|\mathbf{94.36}_{\pm3.53}$ & $46.40|26.22_{\pm23.35}|\mathbf{18.88}_{\pm13.09}$ \\
 & PLCO BRCA & $90.45|92.91_{\pm6.80}|\mathbf{95.65}_{\pm2.68}$ & $91.46|94.15_{\pm5.12}|\mathbf{96.18}_{\pm2.34}$ & $89.10|90.57_{\pm8.79}|\mathbf{94.85}_{\pm2.93}$ & $47.91|36.99_{\pm28.05}|\mathbf{27.02}_{\pm14.76}$ \\
 & PLCO COAD & $97.03|96.00_{\pm3.31}|\mathbf{98.79}_{\pm0.62}$ & $97.17|96.57_{\pm2.72}|\mathbf{98.81}_{\pm0.55}$ & $96.85|94.76_{\pm4.37}|\mathbf{98.55}_{\pm0.69}$ & $16.73|24.26_{\pm20.82}|\mathbf{6.57}_{\pm4.31}$ \\
 & TCGA BRCA & $91.22|92.92_{\pm6.42}|\mathbf{95.49}_{\pm2.58}$ & $90.17|93.78_{\pm5.35}|\mathbf{95.43}_{\pm2.69}$ & $91.88|91.02_{\pm7.44}|\mathbf{95.39}_{\pm2.39}$ & $33.99|36.63_{\pm20.75}|\mathbf{22.64}_{\pm10.28}$ \\
 & TCGA COAD & $93.25|94.41_{\pm3.97}|\mathbf{96.66}_{\pm1.36}$ & $96.36|96.98_{\pm2.20}|\mathbf{98.16}_{\pm0.69}$ & $85.74|87.89_{\pm7.75}|\mathbf{92.99}_{\pm3.16}$ & $36.46|31.81_{\pm19.59}|\mathbf{22.41}_{\pm9.35}$ \\
 & TCGA PAAD & $90.52|96.34_{\pm5.24}|\mathbf{97.23}_{\pm1.76}$ & $97.34|98.91_{\pm1.45}|\mathbf{99.18}_{\pm0.44}$ & $61.79|85.78_{\pm14.59}|\mathbf{87.46}_{\pm8.21}$ & $49.76|18.60_{\pm23.60}|\mathbf{17.26}_{\pm14.59}$ \\
\midrule
\multicolumn{6}{l}{\textbf{Patch-Level}} \\
\midrule
\textit{Very-far OOD} & PLCO IHC & $\mathbf{99.83}|91.79_{\pm10.74}|99.71_{\pm0.32}$ & $\mathbf{99.79}|96.34_{\pm4.77}|99.74_{\pm0.13}$ & $\mathbf{99.44}|80.98_{\pm20.56}|99.00_{\pm1.03}$ & $\mathbf{0.00}|40.00_{\pm42.75}|0.27_{\pm0.40}$ \\
\cmidrule(lr){1-6}
\multirow{4}{*}{\textit{Far OOD}} & CPTAC GBM & $\mathbf{99.67}|93.88_{\pm6.01}|99.44_{\pm0.54}$ & $\mathbf{99.65}|96.38_{\pm3.49}|99.54_{\pm0.32}$ & $\mathbf{99.34}|88.78_{\pm10.80}|98.90_{\pm0.88}$ & $\mathbf{0.76}|33.52_{\pm30.02}|1.97_{\pm3.30}$ \\
 & CPTAC UCEC & $92.90|92.14_{\pm4.92}|\mathbf{94.98}_{\pm3.42}$ & $92.84|92.44_{\pm4.98}|\mathbf{95.05}_{\pm3.28}$ & $93.40|91.75_{\pm5.04}|\mathbf{95.19}_{\pm3.05}$ & $34.37|39.25_{\pm18.83}|\mathbf{23.97}_{\pm16.74}$ \\
 & TCGA GBM & $\mathbf{98.92}|91.36_{\pm8.14}|98.68_{\pm1.29}$ & $\mathbf{98.87}|93.28_{\pm6.10}|98.73_{\pm1.12}$ & $\mathbf{98.80}|88.36_{\pm11.05}|98.38_{\pm1.49}$ & $\mathbf{4.42}|44.90_{\pm35.66}|5.94_{\pm7.72}$ \\
 & TCGA SKCM & $94.77|81.36_{\pm9.29}|\mathbf{94.79}_{\pm2.23}$ & $96.80|89.93_{\pm4.92}|\mathbf{96.84}_{\pm1.21}$ & $\mathbf{91.53}|64.89_{\pm14.95}|90.53_{\pm3.24}$ & $26.00|75.68_{\pm23.12}|\mathbf{24.74}_{\pm13.10}$ \\
\cmidrule(lr){1-6}
\multirow{8}{*}{\textit{Near OOD}} & CPTAC BRCA & $79.73|83.42_{\pm11.08}|\mathbf{86.59}_{\pm6.17}$ & $83.55|89.05_{\pm7.71}|\mathbf{90.05}_{\pm4.70}$ & $77.58|74.38_{\pm14.25}|\mathbf{83.16}_{\pm6.11}$ & $63.32|70.44_{\pm24.27}|\mathbf{53.46}_{\pm13.48}$ \\
 & CPTAC COAD & $68.40|80.77_{\pm10.91}|\mathbf{81.15}_{\pm9.48}$ & $86.09|91.75_{\pm5.61}|\mathbf{92.07}_{\pm4.05}$ & $39.84|56.66_{\pm14.27}|\mathbf{56.80}_{\pm15.00}$ & $96.08|82.07_{\pm17.90}|\mathbf{80.00}_{\pm19.54}$ \\
 & CPTAC PDA & $95.39|89.42_{\pm7.57}|\mathbf{96.49}_{\pm2.71}$ & $96.67|93.32_{\pm4.85}|\mathbf{97.52}_{\pm1.86}$ & $93.67|82.51_{\pm11.49}|\mathbf{94.89}_{\pm3.29}$ & $23.56|49.92_{\pm30.03}|\mathbf{15.79}_{\pm13.02}$ \\
 & PLCO BRCA & $96.32|80.11_{\pm9.22}|\mathbf{96.42}_{\pm1.96}$ & $95.99|84.64_{\pm6.90}|\mathbf{96.48}_{\pm2.10}$ & $\mathbf{96.66}|74.05_{\pm12.56}|96.09_{\pm1.52}$ & $16.86|76.93_{\pm23.97}|\mathbf{15.92}_{\pm9.51}$ \\
 & PLCO COAD & $95.00|81.76_{\pm9.56}|\mathbf{96.31}_{\pm2.20}$ & $94.74|86.65_{\pm6.47}|\mathbf{96.74}_{\pm2.06}$ & $95.22|75.20_{\pm13.77}|\mathbf{95.47}_{\pm1.96}$ & $21.27|73.18_{\pm26.67}|\mathbf{18.96}_{\pm13.95}$ \\
 & TCGA BRCA & $88.04|76.70_{\pm11.16}|\mathbf{90.05}_{\pm5.57}$ & $87.74|81.51_{\pm9.67}|\mathbf{91.00}_{\pm5.03}$ & $88.17|70.12_{\pm12.10}|\mathbf{88.81}_{\pm4.98}$ & $47.66|83.54_{\pm18.97}|\mathbf{43.64}_{\pm19.41}$ \\
 & TCGA COAD & $83.68|85.25_{\pm8.00}|\mathbf{91.69}_{\pm5.69}$ & $91.44|92.14_{\pm4.45}|\mathbf{95.77}_{\pm2.85}$ & $66.96|74.13_{\pm12.27}|\mathbf{82.12}_{\pm10.36}$ & $76.72|61.99_{\pm21.40}|\mathbf{43.47}_{\pm27.81}$ \\
 & TCGA PAAD & $85.75|75.79_{\pm13.05}|\mathbf{88.78}_{\pm6.99}$ & $95.86|93.08_{\pm4.24}|\mathbf{96.87}_{\pm1.97}$ & $57.55|40.27_{\pm17.96}|\mathbf{63.89}_{\pm14.26}$ & $63.16|82.90_{\pm21.11}|\mathbf{52.65}_{\pm25.55}$ \\
\bottomrule
\end{tabular}}
\end{table*}
\begin{table*}[h!]
\centering
\caption{ \footnotesize \textbf{Comparison with SOTA OOD methods.} ID:CPTAC-LUAD/LUSC. Performance scores: AUROC $|$ AUIN $|$ AUOUT $|$ FPR95.}
\label{tab:sota_comparison_titan_conch}
\begingroup
\small
\setlength{\tabcolsep}{3pt}
\renewcommand{\arraystretch}{0.96}
\resizebox{\textwidth}{!}{%
\begin{tabular}{@{}l c c c c c c@{}}
\toprule
\textbf{Method} & \textbf{CPTAC\_PDA} & \textbf{CPTAC\_GBM} & \textbf{PLCO\_IHC} & \textbf{PLCO\_COAD} & \textbf{TCGA\_BRCA} & \textbf{TCGA\_PAAD}  \\
\midrule
\multicolumn{5}{l}{\textbf{Slide-level}} \\ \midrule
msp~\cite{hendrycks2017baseline} & 80.89 $|$ 87.27 $|$ 69.64 $|$ 78.96 & 77.48 $|$ 86.24 $|$ 63.54 $|$ 84.19 & 86.01 $|$ 93.21 $|$ 68.77 $|$ 76.50 & 77.89 $|$ 81.51 $|$ 73.58 $|$ 85.34 & 78.08 $|$ 80.48 $|$ 74.46 $|$ 82.31 & 80.66 $|$ 94.38 $|$ 46.58 $|$ 79.43 \\
odin~\cite{liang2018enhancing} & 78.07 $|$ 83.64 $|$ 69.00 $|$ 73.92 & 71.30 $|$ 80.69 $|$ 53.75 $|$ 92.00 & 83.54 $|$ 91.30 $|$ 66.30 $|$ 75.58 & 71.12 $|$ 75.94 $|$ 61.77 $|$ 96.81 & 74.33 $|$ 76.32 $|$ 67.96 $|$ 92.32 & 72.28 $|$ 90.93 $|$ 35.87 $|$ 84.69  \\
maha~\cite{lee2018simple}  & 91.08 $|$ 94.37 $|$ 85.20 $|$ 51.08 & 94.88 $|$ 96.83 $|$ 91.48 $|$ 33.33 & 91.47 $|$ 95.66 $|$ 82.29 $|$ 52.53 & 94.85 $|$ 95.11 $|$ 94.44 $|$ 30.61 & 93.63 $|$ 94.31 $|$ 92.63 $|$ 41.56 & 95.28 $|$ 98.66 $|$ 81.85 $|$ 33.01  \\
energy~\cite{liu2020energy} & 80.35 $|$ 87.07 $|$ 66.80 $|$ 78.42 & 76.57 $|$ 85.80 $|$ 59.60 $|$ 85.52 & 85.46 $|$ 92.96 $|$ 66.71 $|$ 75.81 & 77.45 $|$ 81.11 $|$ 72.28 $|$ 84.28 & 77.18 $|$ 79.76 $|$ 72.18 $|$ 82.77 & 80.40 $|$ 94.37 $|$ 42.56 $|$ 80.86  \\
gradnorm~\cite{b14} & 80.92 $|$ 86.95 $|$ 70.35 $|$ 75.18 & 68.69 $|$ 79.44 $|$ 56.22 $|$ 83.81 & 82.63 $|$ 90.87 $|$ 65.68 $|$ 74.88 & 70.77 $|$ 74.10 $|$ 67.95 $|$ 85.82 & 76.61 $|$ 78.83 $|$ 73.37 $|$ 79.98 & 78.26 $|$ 93.21 $|$ 46.03 $|$ 76.08 \\
knn~\cite{b31} & 80.86 $|$ 87.63 $|$ 68.37 $|$ 80.76 & 78.35 $|$ 87.45 $|$ 60.49 $|$ 88.57 & 85.90 $|$ 93.34 $|$ 66.54 $|$ 80.88 & 79.61 $|$ 83.68 $|$ 72.12 $|$ 88.06 & 79.16 $|$ 82.16 $|$ 73.79 $|$ 84.75 & 78.14 $|$ 93.72 $|$ 39.54 $|$ 87.56 \\
react~\cite{b28} & 72.70 $|$ 81.94 $|$ 57.11 $|$ 89.93 & 73.21 $|$ 83.77 $|$ 53.67 $|$ 93.33 & 78.83 $|$ 89.64 $|$ 56.63 $|$ 88.02 & 71.53 $|$ 75.54 $|$ 64.06 $|$ 92.55 & 68.73 $|$ 71.97 $|$ 63.17 $|$ 92.55 & 77.83 $|$ 93.52 $|$ 38.52 $|$ 85.17 \\
vim~\cite{wang2022vim} & 87.43 $|$ 91.59 $|$ 80.70 $|$ 55.04 & 86.38 $|$ 91.54 $|$ 78.54 $|$ 58.86 & 91.47 $|$ 95.68 $|$ 81.92 $|$ 46.08 & 86.57 $|$ 87.54 $|$ 86.15 $|$ 53.90 & 85.47 $|$ 86.39 $|$ 84.63 $|$ 59.95 & 89.16 $|$ 96.83 $|$ 67.43 $|$ 45.93  \\
maxlogit~\cite{hendrycks2022scaling} & 80.51 $|$ 87.11 $|$ 68.06 $|$ 77.52 & 76.79 $|$ 85.86 $|$ 61.29 $|$ 83.62 & 85.55 $|$ 92.99 $|$ 67.43 $|$ 75.58 & 77.50 $|$ 81.13 $|$ 72.90 $|$ 84.40 & 77.34 $|$ 79.81 $|$ 73.04 $|$ 81.72 & 80.55 $|$ 94.40 $|$ 44.57 $|$ 78.95  \\
nci~\cite{liu2025detecting} & 80.74 $|$ 86.83 $|$ 68.94 $|$ 78.78 & 70.54 $|$ 80.49 $|$ 55.31 $|$ 88.76 & 83.88 $|$ 91.61 $|$ 65.92 $|$ 78.57 & 74.01 $|$ 77.08 $|$ 69.52 $|$ 87.83 & 76.28 $|$ 77.04 $|$ 72.16 $|$ 84.40 & 79.58 $|$ 94.00 $|$ 43.75 $|$ 81.82  \\
scale~\cite{xu2024scaling} & 77.01 $|$ 78.51 $|$ 68.08 $|$ 78.24 & 71.69 $|$ 75.42 $|$ 60.96 $|$ 84.00 & 84.11 $|$ 90.49 $|$ 67.87 $|$ 76.04 & 71.79 $|$ 67.20 $|$ 71.27 $|$ 84.99 & 73.59 $|$ 66.86 $|$ 72.29 $|$ 82.07 & 75.89 $|$ 90.81 $|$ 44.60 $|$ 79.43  \\
fdbd~\cite{fdbd} & 80.89 $|$ 85.83 $|$ 70.83 $|$ 74.64 & 69.89 $|$ 76.46 $|$ 57.05 $|$ 85.52 & 84.32 $|$ 91.92 $|$ 66.98 $|$ 75.58 & 72.05 $|$ 72.32 $|$ 68.65 $|$ 86.29 & 75.85 $|$ 72.88 $|$ 73.41 $|$ 81.14 & 80.18 $|$ 94.07 $|$ 47.51 $|$ 77.03 \\
oe~\cite{b8} & 83.61 $|$ 90.40 $|$ 69.45 $|$ 80.22 & 80.09 $|$ 88.52 $|$ 63.58 $|$ 84.76 & 89.13 $|$ 94.98 $|$ 72.39 $|$ 71.43 & 79.21 $|$ 83.95 $|$ 71.89 $|$ 86.64 & 85.58 $|$ 88.74 $|$ 80.35 $|$ 77.18 & 87.69 $|$ 96.77 $|$ 54.24 $|$ 77.03  \\
cider~\cite{ming2023cider} & 79.30 $|$ 86.78 $|$ 66.67 $|$ 78.24 & 87.73 $|$ 92.42 $|$ 78.00 $|$ 62.29 & 69.87 $|$ 83.53 $|$ 47.50 $|$ 91.01 & 81.35 $|$ 83.84 $|$ 78.07 $|$ 73.64 & 83.90 $|$ 85.90 $|$ 81.21 $|$ 68.92 & 80.21 $|$ 94.24 $|$ 44.37 $|$ 77.99  \\
lnorm~\cite{wei2022mitigating}  & 78.46 $|$ 85.04 $|$ 64.70 $|$ 84.17 & 74.34 $|$ 81.69 $|$ 58.90 $|$ 86.29 & 77.92 $|$ 87.81 $|$ 58.52 $|$ 85.48 & 59.34 $|$ 57.01 $|$ 59.50 $|$ 91.02 & 75.93 $|$ 73.05 $|$ 73.25 $|$ 83.12 & 77.81 $|$ 92.50 $|$ 43.01 $|$ 83.25  \\

\midrule
\textbf{ZIO} & \textbf{96.57 $|$ 97.66 $|$ 94.36 $|$ 18.88} & \textbf{98.89 $|$ 99.19 $|$ 98.01 $|$ 4.93} & \textbf{99.96 $|$ 99.85 $|$ 99.70 $|$ 0.11} & \textbf{98.79 $|$ 98.81 $|$ 98.55 $|$ 6.57} & \textbf{95.49 $|$ 95.43 $|$ 95.39 $|$ 22.64} & \textbf{97.23 $|$ 99.18 $|$ 87.46 $|$ 17.26}   \\

\midrule \multicolumn{5}{l}{\textbf{Patch-level}} \\ \midrule
msp~\cite{hendrycks2017baseline}        & 64.44 $|$ 76.00 $|$ 45.50 $|$ 99.10 & 69.71 $|$ 83.03 $|$ 48.57 $|$ 96.00 & 69.31 $|$ 85.52 $|$ 42.39 $|$ 100.0 & 68.95 $|$ 72.76 $|$ 64.40 $|$ 89.48 & 61.20 $|$ 66.15 $|$ 53.83 $|$ 98.84 & 48.46 $|$ 80.77 $|$ 18.37 $|$ 97.61  \\
odin~\cite{liang2018enhancing}       & 65.43 $|$ 76.19 $|$ 46.73 $|$ 100.0 & 62.03 $|$ 79.11 $|$ 41.66 $|$ 100.0 & 72.50 $|$ 86.63 $|$ 45.29 $|$ 100.0 & 60.21 $|$ 67.75 $|$ 53.86 $|$ 100.0 & 59.50 $|$ 64.25 $|$ 53.45 $|$ 100.0 & 45.19 $|$ 80.12 $|$ 17.19 $|$ 100.0 \\
maha~\cite{lee2018simple}        & 61.84 $|$ 72.64 $|$ 51.35 $|$ 85.61 & 73.80 $|$ 81.88 $|$ 64.48 $|$ 71.81 & 54.86 $|$ 76.83 $|$ 36.83 $|$ 93.78 & 84.05 $|$ 83.11 $|$ 84.47 $|$ 47.99 & 71.28 $|$ 71.92 $|$ 70.49 $|$ 78.11 & 83.75 $|$ 94.75 $|$ 52.28 $|$ 63.16  \\

energy~\cite{liu2020energy}     & 60.57 $|$ 74.09 $|$ 42.57 $|$ 100.0 & 64.75 $|$ 80.55 $|$ 43.56 $|$ 100.0 & 66.93 $|$ 84.36 $|$ 40.67 $|$ 100.0 & 63.50 $|$ 69.63 $|$ 55.86 $|$ 100.0 & 57.93 $|$ 63.84 $|$ 51.47 $|$ 100.0 & 44.43 $|$ 79.08 $|$ 17.01 $|$ 100.0  \\
gradnorm~\cite{b14}   & 69.76 $|$ 80.81 $|$ 49.84 $|$ 99.10 & 73.70 $|$ 85.10 $|$ 52.33 $|$ 95.81 & 75.27 $|$ 88.48 $|$ 47.52 $|$ 100.0 & 71.48 $|$ 75.87 $|$ 67.38 $|$ 84.16 & 65.64 $|$ 69.39 $|$ 57.35 $|$ 98.60 & 54.13 $|$ 84.64 $|$ 20.23 $|$ 97.61  \\
knn~\cite{b31}        & 49.31 $|$ 61.91 $|$ 38.58 $|$ 98.38 & 58.82 $|$ 68.10 $|$ 45.21 $|$ 93.33 & 43.37 $|$ 67.38 $|$ 29.04 $|$ 99.54 & 72.94 $|$ 74.19 $|$ 67.20 $|$ 91.61 & 59.41 $|$ 60.72 $|$ 57.67 $|$ 93.95 & 70.56 $|$ 90.03 $|$ 30.29 $|$ 97.13 \\
react~\cite{b28}      & 61.78 $|$ 69.23 $|$ 50.01 $|$ 90.11 & 45.97 $|$ 58.00 $|$ 35.83 $|$ 97.71 & 63.86 $|$ 79.69 $|$ 44.12 $|$ 90.78 & 46.01 $|$ 49.06 $|$ 45.28 $|$ 98.82 & 56.17 $|$ 52.73 $|$ 58.74 $|$ 88.47 & 53.16 $|$ 82.58 $|$ 19.78 $|$ 98.09  \\
vim~\cite{wang2022vim}        & 72.07 $|$ 81.58 $|$ 58.17 $|$ 86.33 & 83.20 $|$ 89.55 $|$ 74.38 $|$ 64.19 & 71.49 $|$ 86.66 $|$ 46.22 $|$ 94.24 & 89.75 $|$ 90.33 $|$ 89.25 $|$ 47.87 & 78.82 $|$ 81.38 $|$ 74.98 $|$ 79.86 & 80.13 $|$ 93.56 $|$ 46.01 $|$ 78.47  \\
maxlogit~\cite{hendrycks2022scaling}   & 61.52 $|$ 74.48 $|$ 43.24 $|$ 100.0 & 66.08 $|$ 81.19 $|$ 44.59 $|$ 99.62 & 67.09 $|$ 84.48 $|$ 40.81 $|$ 100.0 & 65.08 $|$ 70.34 $|$ 57.80 $|$ 99.88 & 58.47 $|$ 64.17 $|$ 51.78 $|$ 100.0 & 45.24 $|$ 79.29 $|$ 17.24 $|$ 100.0  \\
nci~\cite{liu2025detecting}        & 71.31 $|$ 78.96 $|$ 52.16 $|$ 98.38 & 74.02 $|$ 81.97 $|$ 54.93 $|$ 94.48 & 77.75 $|$ 88.66 $|$ 50.54 $|$ 99.77 & 70.89 $|$ 69.36 $|$ 69.49 $|$ 81.32 & 65.93 $|$ 62.66 $|$ 59.56 $|$ 97.79 & 52.94 $|$ 82.74 $|$ 20.14 $|$ 98.56  \\
scale~\cite{xu2024scaling}      & 64.73 $|$ 75.83 $|$ 45.72 $|$ 99.10 & 69.99 $|$ 83.10 $|$ 48.77 $|$ 96.00 & 69.70 $|$ 85.73 $|$ 42.66 $|$ 100.0 & 69.07 $|$ 72.43 $|$ 64.51 $|$ 89.72 & 61.39 $|$ 64.91 $|$ 53.99 $|$ 98.84 & 48.82 $|$ 81.11 $|$ 18.50 $|$ 97.61  \\
fdbd~\cite{fdbd}       & 63.24 $|$ 72.39 $|$ 44.95 $|$ 99.46 & 67.45 $|$ 81.38 $|$ 46.06 $|$ 97.71 & 69.47 $|$ 84.23 $|$ 42.76 $|$ 100.0 & 64.31 $|$ 66.70 $|$ 59.88 $|$ 95.86 & 58.24 $|$ 61.19 $|$ 52.12 $|$ 99.53 & 44.58 $|$ 78.04 $|$ 17.23 $|$ 98.09  \\
oe~\cite{b8}         & 78.37 $|$ 86.64 $|$ 61.89 $|$ 88.85 & 81.87 $|$ 90.07 $|$ 64.01 $|$ 87.81 & 90.81 $|$ 95.48 $|$ 77.12 $|$ 61.29 & 78.30 $|$ 84.05 $|$ 69.80 $|$ 90.66 & 80.16 $|$ 83.74 $|$ 73.48 $|$ 89.06 & 68.74 $|$ 91.12 $|$ 27.77 $|$ 95.69  \\
cider~\cite{ming2023cider}      & 62.79 $|$ 75.86 $|$ 48.69 $|$ 92.45 & 58.16 $|$ 74.10 $|$ 43.95 $|$ 93.71 & 55.80 $|$ 78.28 $|$ 34.56 $|$ 97.70 & 63.26 $|$ 66.77 $|$ 61.88 $|$ 89.48 & 63.54 $|$ 68.70 $|$ 60.46 $|$ 92.90 & 68.78 $|$ 89.41 $|$ 32.94 $|$ 87.08  \\
lnorm~\cite{wei2022mitigating}      & 68.93 $|$ 80.73 $|$ 49.56 $|$ 98.02 & 81.33 $|$ 89.82 $|$ 62.83 $|$ 90.10 & 66.97 $|$ 85.45 $|$ 40.28 $|$ 99.77 & 79.53 $|$ 83.57 $|$ 72.94 $|$ 87.59 & 68.62 $|$ 75.61 $|$ 61.57 $|$ 94.30 & 59.41 $|$ 86.79 $|$ 22.62 $|$ 95.69  \\ \midrule
\textbf{ZIO} & \textbf{96.49 $|$ 97.52 $|$ 94.89 $|$ 15.79} & \textbf{99.44 $|$ 99.54 $|$ 98.90 $|$ 1.97} & \textbf{99.71 $|$ 99.74 $|$ 99.00 $|$ 0.27} & \textbf{96.31 $|$ 96.74 $|$ 95.47 $|$ 18.96} & \textbf{90.05 $|$ 91.00 $|$ 88.81 $|$ 43.64} & \textbf{88.78 $|$ 96.87 $|$ 63.89 $|$ 52.65} \\
\bottomrule
\end{tabular}}%
\endgroup
\end{table*}


\vspace{1.5mm}
\noindent\textbf{(i) ZIO outperforms unimodal ZS and OS baselines:} \Cref{tab:titan_conch_zs_os_zio_ood_grouped}
shows the performance of the ZIO model, in comparison with the unimodal ZS and OS baselines across \textit{very-far}, \textit{far}, and \textit{near-OOD} cases. Both ZS and OS models reach relatively high performance, however neither consistently dominates across all cancer types and datasets. This suggests that the visual and text prototypes have complementary properties. ZIO efficiently integrates this complementary information from both modalities, leading to consistent improvement over both unimodal baselines.

\vspace{1.5mm}
\noindent\textbf{(ii) ZIO generalizes across cancer types and external cohorts:}
\Cref{tab:titan_conch_zs_os_zio_ood_grouped} further demonstrates that ZIO maintains robust performance across external cohorts and diverse malignancies. For example, ZIO achieves an AUROC of 0.95 for BRCA across TCGA, PLCO, and CPTAC cohorts, indicating strong generalization to variations in data distribution arising from different institutions and data preparation protocols. At the same time, ZIO also demonstrates robust performance across various malignancies and domain shifts.

\vspace{1.5mm}
\noindent\textbf{(iii) ZIO outperforms SOTA OOD methods:} \Cref{tab:sota_comparison_titan_conch} reports results for 15 representative OOD methods evaluated on six datasets (restricted to two per consortium due to space constraints), while \cref{fig:40sota} shows comparison with all 40 methods, averaged across all 13 datasets presented in \cref{tab:titan_conch_zs_os_zio_ood_grouped}. Across all evaluations, ZIO consistently outperforms all 40 SOTA methods, for both slide- and patch-level formulations. Since all methods use exactly the same FM embeddings, these results demonstrate the effectiveness of the ZIO model rather than the benefits of the FMs alone. Moreover, to isolate the added value of the multimodal prototype learning, ZIO uses a simple MaxLogit scoring function, which can be easily replaced with more advanced methods.

\begin{figure*}[t]
\centerline{\includegraphics[width=0.72\textwidth]{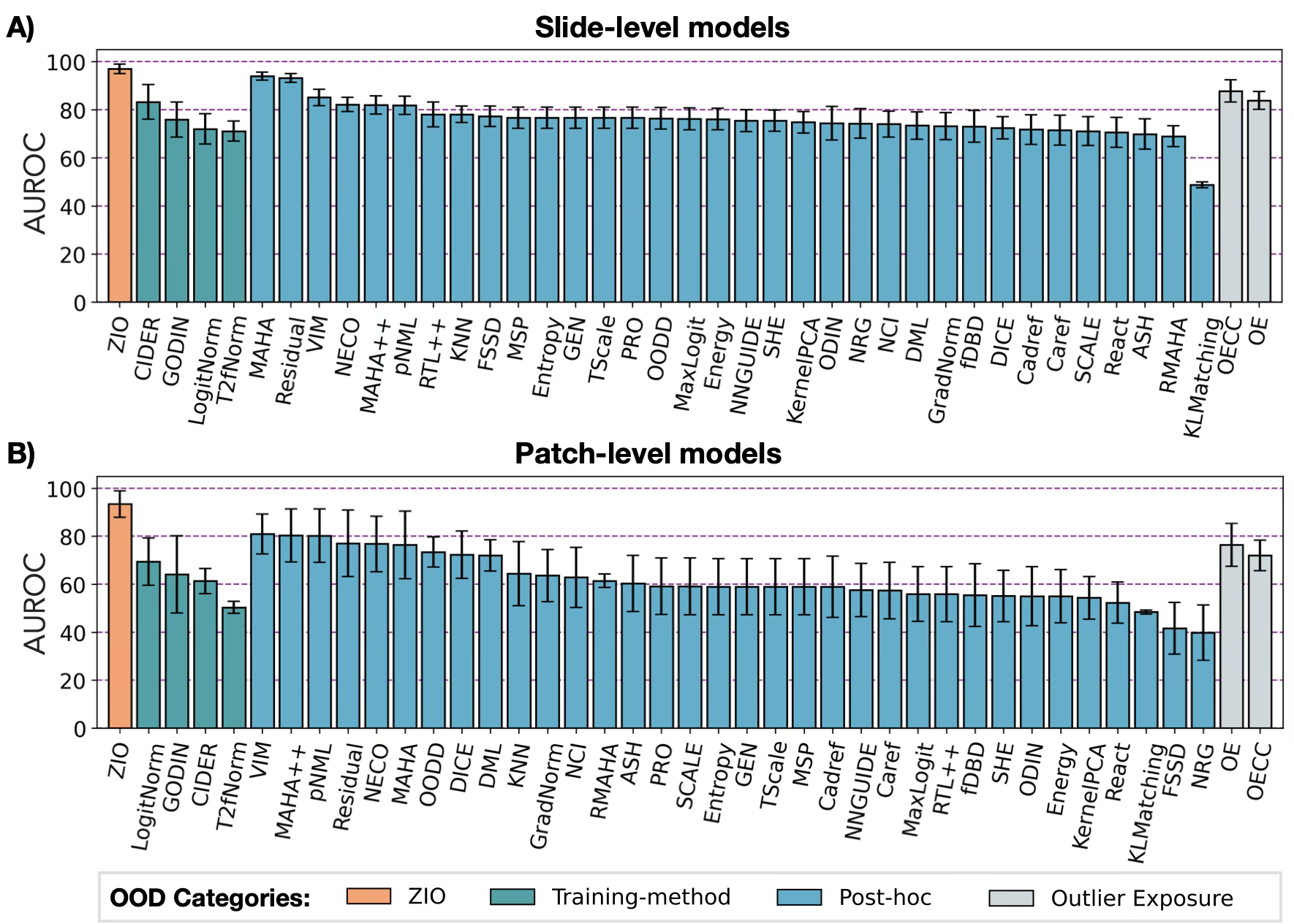}}
\caption{\textbf{Comparison with SOTA methods.} AUROC (mean and std) computed across 13 datasets comparing 40 SOTA OOD detectors
with ZIO for (A) slide- and (B) patch-level representations. The 40 OOD baselines include: ASH~\cite{djurisic2022extremely}, Cadref~\cite{ling2025cadref}, Caref~\cite{ling2025cadref}, CIDER~\cite{ming2023cider}, DICE~\cite{sun2022dice}, DML~\cite{zhang2023decoupling}, Energy~\cite{liu2020energy}, Entropy~\cite{ren2019likelihood}, fDBD~\cite{fdbd}, fssd~\cite{b6}, GEN~\cite{liu2023gen}, GODIN~\cite{hsu2020generalized}, GradNorm~\cite{b14}, 
KernelPCA~\cite{fang2024kernel}, 
KLMatching~\cite{hendrycks2022scaling}, 
KNN~\cite{b31}, 
LogitNorm~\cite{wei2022mitigating}, 
MAHA~\cite{lee2018simple}, MAHA++~\cite{mueller2025mahalanobis++}, 
MaxLogit~\cite{hendrycks2022scaling}, 
MSP~\cite{hendrycks2017baseline}, 
NCI~\cite{liu2025detecting}, 
NECO~\cite{ammar2023neco}, 
NNGUIDE~\cite{park2023nearest}, 
NRG~\cite{kim2023neural}, 
ODIN~\cite{liang2018enhancing}, 
OE~\cite{b8}, OECC~\cite{papadopoulos2021outlier}, 
OODD~\cite{yang2025oodd}, 
pNML~\cite{bibas2021single}, 
PRO~\cite{chen2025leveraging}, 
React~\cite{b28}, 
Residual~\cite{wang2022vim}, 
RMAHA~\cite{ren2021simple}, 
RTL++~\cite{fan2024test}, 
SCALE~\cite{xu2024scaling}, SHE~\cite{zhang2023outofdistribution}, T2fNorm~\cite{regmi2024t2fnorm}, TScale~\cite{guo2017calibration}, and VIM~\cite{wang2022vim}.}
\label{fig:40sota}
\vspace{-3mm}
\end{figure*}

\vspace{-3mm}
\subsection{Within-Organ (very-near-OOD) Detection}
While the previous experiments showed generalization across cancer types, here we analyze clinically more relevant and challenging scenarios. As before, we use NSCLC subtyping as the ID-task, while the OOD cases include other pathologies in the lung tissue. We evaluate three \textit{very near-within-organ OOD} cases: \textit{(i) GTEx-near1: Non-malignant findings} such as calcification, atelectasis or hypertrophy in lung tissue, \textit{(ii) GTEx-near2: Non-malignant cancer mimickers,} which include benign lung pathologies that share features with lung cancer, such as reactive or inflammatory tissue, fibrosis, or edema, \textit{(iii) PLCO RareLung:} includes rare lung cancer types, such as small-cell lung carcinoma or large-cell lung carcinoma. \Cref{tab:within-organ_nearood_comparison} shows the performance of ZIO in comparison with top five OOD detectors (based on their average performance). ZIO reached AUROC over 0.97 for the non-malignant cases, while the performance on the rare cancers is 0.87, suggesting the increased complexity of rare-lung cancer detection. Nonetheless, ZIO outperformed all SOTA OOD methods by 10-20\%.

\vspace{-2mm}
\begin{table}[h]
\centering
\caption{\scriptsize\textbf{Very-near within organ OOD detection (slide-level).} \\ ID: TCGA-LUAD/LUSC. Values are AUROC}
\label{tab:within-organ_nearood_comparison}
\scriptsize
\setlength{\tabcolsep}{3pt}
\begin{tabular}{lccc}
\toprule
Method &(\textit{i}) GTEx-near1  &(\textit{ii}) GTEx-near2  & (\textit{iii}) PLCO-RareLung\\
\midrule
ZIO      & \textbf{97.59} & \textbf{97.34} & \textbf{87.04} \\
MAHA & 72.33 & 74.41 & 79.68 \\
Residual & 70.99 & 70.63 & 76.56 \\
OECC & 49.73 & 55.10 & 74.08 \\
VIM & 73.79 & 79.40 & 79.21 \\
OE & 75.15 & 81.77 & 69.44 \\
\bottomrule
\end{tabular}
\vspace{-1mm}
\end{table}

\setlength{\dbltextfloatsep}{6pt plus 2pt minus 2pt} 
\begin{table*}[h!]
\centering 
\scriptsize 
\setlength{\tabcolsep}{3pt}
\renewcommand{\arraystretch}{0.92}
\caption{\footnotesize\textbf{Performance on rare brain cancers}. All metrics are shown as $\textit{ZS} | \textit{OS} | \textit{ZIO}$. (ID: EBRAINS-RB$_0$) } 
\label{tab:rare_brain_lung_titan_conch}
\resizebox{0.75\textwidth}{!}{
\begin{tabular}{@{}l c c c c@{}}
\toprule
OOD Dataset & \textbf{AUROC}& \textbf{AUIN} & \textbf{AUOUT} & \textbf{FPR95} \\
\midrule
\multicolumn{5}{l}{\textbf{Slide-Level }} \\
\midrule
RB$_1$ & $88.99|96.65_{\pm1.54}|\mathbf{97.01}_{\pm0.92}$ & $96.51|99.00_{\pm0.43}|\mathbf{99.11}_{\pm0.26}$ & $78.02|89.71_{\pm3.75}|\mathbf{90.96}_{\pm1.68}$ & $32.07|15.42_{\pm6.36}|\mathbf{15.19}_{\pm3.87}$ \\
RB$_2$ & $88.94|96.17_{\pm2.59}|\mathbf{96.96}_{\pm1.58}$ & $91.80|97.06_{\pm1.77}|\mathbf{97.79}_{\pm1.03}$ & $82.89|93.70_{\pm4.40}|\mathbf{94.50}_{\pm2.86}$ & $54.04|18.63_{\pm15.68}|\mathbf{16.17}_{\pm11.96}$ \\
RB$_3$ & $87.22|89.90_{\pm2.66}|\mathbf{92.55}_{\pm1.30}$ & $69.43|72.18_{\pm6.99}|\mathbf{78.51}_{\pm2.90}$ & $95.64|96.62_{\pm0.82}|\mathbf{97.51}_{\pm0.44}$ & $46.08|35.36_{\pm5.87}|\mathbf{28.01}_{\pm5.39}$ \\
\midrule
\multicolumn{5}{l}{\textbf{Patch-Level }} \\
\midrule
RB$_1$ & $81.30|79.47_{\pm5.01}|\mathbf{86.33}_{\pm1.70}$ & $92.69|93.56_{\pm2.00}|\mathbf{95.70}_{\pm0.79}$ & $63.85|49.39_{\pm9.54}|\mathbf{68.04}_{\pm2.66}$ & $47.56|66.60_{\pm11.34}|\mathbf{44.60}_{\pm3.69}$ \\
RB$_2$ & $79.96|76.01_{\pm6.23}|\mathbf{88.70}_{\pm2.80}$ & $84.69|82.01_{\pm5.83}|\mathbf{91.51}_{\pm2.44}$ & $72.36|66.21_{\pm6.68}|\mathbf{82.53}_{\pm3.48}$ & $75.50|82.26_{\pm6.63}|\mathbf{58.08}_{\pm9.51}$ \\
RB$_3$ & $73.58|62.55_{\pm5.47}|\mathbf{74.93}_{\pm2.65}$ & $46.24|32.91_{\pm7.55}|\mathbf{46.33}_{\pm5.26}$ & $89.48|84.50_{\pm2.47}|\mathbf{90.48}_{\pm0.99}$ & $77.38|84.73_{\pm4.18}|\mathbf{71.16}_{\pm3.54}$ \\
\bottomrule
\end{tabular}}
\end{table*}

\begin{table*}[h]
\centering
\scriptsize
\setlength{\tabcolsep}{3pt}
\renewcommand{\arraystretch}{0.92}
\caption{\footnotesize\textbf{Performance on IHC stains.} ID: PLCO-IHC, OOD: H\&E-stained tissue. All metrics show $\textit{ZS} | \textit{OS} | \textit{ZIO}$.}
\label{tab:zs_os_zio_ihc_titan_conch}
\resizebox{0.75\textwidth}{!}{
\begin{tabular}{@{}l c c c c@{}}
\toprule
\textbf{OOD dataset} & \textbf{AUROC} & \textbf{AUIN} & \textbf{AUOUT} & \textbf{FPR95} \\
\midrule
\multicolumn{5}{l}{\textbf{Slide-Level}} \\
\midrule
EBRAINS (ALL) & $14.73|\mathbf{99.97}_{\pm0.06}|99.61_{\pm0.80}$ & $7.78|\mathbf{99.65}_{\pm0.25}|97.79_{\pm5.45}$ & $76.17|\mathbf{99.96}_{\pm0.01}|99.90_{\pm0.13}$ & $96.01|\mathbf{0.04}_{\pm0.14}|1.50_{\pm3.45}$ \\
CPTAC GBM & $76.64|\mathbf{99.91}_{\pm0.21}|99.85_{\pm0.42}$ & $59.49|\mathbf{99.65}_{\pm0.29}|99.49_{\pm0.92}$ & $85.92|\mathbf{99.74}_{\pm0.15}|99.70_{\pm0.26}$ & $42.10|\mathbf{0.29}_{\pm1.09}|0.51_{\pm1.46}$ \\
TCGA GBM & $79.04|\mathbf{99.98}_{\pm0.03}|\mathbf{99.98}_{\pm0.07}$ & $51.16|\mathbf{99.74}_{\pm0.07}|99.71_{\pm0.18}$ & $90.88|\mathbf{99.88}_{\pm0.02}|99.87_{\pm0.03}$ & $40.63|\mathbf{0.05}_{\pm0.21}|0.07_{\pm0.24}$ \\
\midrule
\multicolumn{5}{l}{\textbf{Patch-Level}} \\
\midrule
EBRAINS (ALL) & $91.91|\mathbf{99.89}_{\pm0.05}|99.75_{\pm1.05}$ & $66.04|\mathbf{99.27}_{\pm0.24}|98.43_{\pm5.22}$ & $98.55|\mathbf{99.95}_{\pm0.01}|99.92_{\pm0.17}$ & $31.54|\mathbf{0.42}_{\pm0.51}|0.89_{\pm3.91}$ \\
CPTAC GBM & $95.15|\mathbf{99.99}_{\pm0.01}|99.45_{\pm0.59}$ & $89.79|\mathbf{99.76}_{\pm0.02}|98.75_{\pm1.48}$ & $96.66|\mathbf{99.80}_{\pm0.01}|99.46_{\pm0.33}$ & $13.71|\mathbf{0.00}_{\pm0.00}|1.78_{\pm1.16}$ \\
TCGA GBM & $94.56|\mathbf{99.98}_{\pm0.01}|99.92_{\pm0.20}$ & $89.54|\mathbf{99.74}_{\pm0.02}|99.60_{\pm0.48}$ & $96.91|\mathbf{99.87}_{\pm0.01}|99.85_{\pm0.09}$ & $25.66|\mathbf{0.00}_{\pm0.02}|0.26_{\pm0.77}$ \\
\bottomrule
\end{tabular}}
\end{table*}

\vspace{1mm}
\subsection{Very-Near-OOD Detection for Rare Brain Cancers}
\label{sec:exp2}
The previous experiments consider common cancer types as ID samples. Since ZIO is a training-free method, it can be applied as a safeguard also in low-data regimes. To illustrate this, here we consider rare brain cancer types as ID classes, while the OOD set comprises both common and rare brain cancers. This setting is particularly challenging due to the strong morphological and biological similarity between ID and OOD samples. It mimics a realistic clinical setting, where for instance, an AI model trained for treatment response predictions in meningiomas should abstain from predictions when presented with a GBM sample. We use the EBRAINS~\cite{amunts2016human} cohort, which consists of 125 distinct brain cancer types, and divide it into four histologically distinct datasets (based on neuro-pathologist assessment): \textit{RB$_0$} contains 7 rare cancer types (pilocytic astrocytoma, ganglioglioma, ependymoma, adamantinomatous craniopharyngioma, pituitary adenoma, schwannoma, haemangioblastoma) and serves as the ID set. The remaining data are divided into three OOD configurations: \textit{RB$_1$} consists of 14 cancer subtypes comprising brain lymphomas, hematopoietic, and histiocytic disorders; \textit{RB$_2$} includes 15 meningioma subtypes; and \textit{RB$_3$} covers all 118 cancer types not present in the ID set ( including also \textit{RB$_1$} and \textit{RB$_2$}). Some cancer subtypes (classes) are represented by very limited sample sizes (e.g. 12 WSIs per class), reflecting the data scarcity of rare diseases. Since many conventional OOD methods require a training of an ID classifier on sufficiently large datasets, they cannot be applied in this low-data setting. The performance of ZIO and the unimodal baselines is reported in \Cref{tab:rare_brain_lung_titan_conch}. While all prototype-based approaches reached strong performance, ZIO outperformed both unimodal models. The performance at the patch level is, however, lower compared to the slide level. One possible explanation is that patch-level representations lack the broader tissue context that might be required to identify the subtle difference between ID and OOD brain cancer subtypes. This experiment demonstrates the feasibility of the ZIO deployment under realistic, data-constrained, and semantically challenging very near-OOD conditions.

\vspace{-1mm}
\subsection{OOD Detection under Stain Shift}
\label{sec:exp3}
The previous experiments evaluated H\&E-stained WSIs as ID samples. Here, we instead consider IHC-stained slides as ID cases and treat H\&E samples as OOD. This setup evaluates robustness to covariate shifts induced by different staining mechanisms: IHC highlights specific protein expression patterns, whereas H\&E emphasizes tissue morphology. Results in \cref{tab:zs_os_zio_ihc_titan_conch} show that the text-only prototypes achieve variable performance and struggle under this type of domain shift. In contrast, both the image-only and multimodal prototypes achieve near-perfect performance. These findings reinforce our earlier observations that unimodal approaches may be sensitive to certain distribution shifts, whereas the multimodal formulation provides a more stable and reliable solution.

\vspace{-1mm}
\subsection{Ablation Studies}
\label{sec:ablations}

\noindent\textbf{Data efficiency.}
We assess the data efficiency of ZIO with respect to the number of WSIs used to construct visual prototypes. We compare the performance of ZIO using one vs.\ multiple image prototypes. \Cref{tab:ablation-one-vs-multishot-auroc-cptac-patch-level} shows the ablation study for the patch-level models on common cancer types, while \cref{tab:ablation-one-vs-multishot-auroc-ebrains} shows the results for slide-level models on rare brain cancers. These findings show that multi-shot variants provide only marginal improvements over the one-shot (default ZIO) formulation, demonstrating the model data efficiency.
 
\vspace{1mm}
\noindent\textbf{Top-$j$ pooling.} We analyze the effect of the top-$j$ selection in the patch-level models. \Cref{tab:topj-ablation} shows ZIO performance with $j \in \{1, 5, 10, 50, 100\}$. The results confirm that $j=10$ (the default value) achieves the best performance, outperforming both smaller $j$ values (which might not sufficiently capture the disease representation) and larger values (which may include diagnostically non-relevant tissue).
\begin{table}[t!]
\centering
\caption{\footnotesize\textbf{Ablation study for ZS-informed one vs.\ multi-shot with patch-level FM}. ID: CPTAC-LUAD/LUSC. Values are AUROC.}
\label{tab:ablation-one-vs-multishot-auroc-cptac-patch-level}
\resizebox{\columnwidth}{!}{%
\setlength{\tabcolsep}{3pt}
\scriptsize
\begin{tabular}{llcccc}
\toprule
\textbf{Cohort} & \textbf{OOD Dataset} & \textbf{1-shot} & \textbf{5-shot} & \textbf{10-shot} & \textbf{50-shot} \\
\midrule
CPTAC & BRCA & $86.59$ & $86.02_{\pm3.08}$ & $86.95_{\pm3.08}$ & $87.11_{\pm1.69}$ \\
      & COAD & $81.15$ & $82.45_{\pm7.04}$ & $84.37_{\pm4.98}$ & $85.03_{\pm2.64}$ \\
      & GBM  & $99.44$ & $99.77_{\pm0.08}$ & $99.82_{\pm0.05}$ & $99.84_{\pm0.01}$ \\
      & PDA  & $96.49$ & $97.90_{\pm1.38}$ & $98.19_{\pm0.99}$ & $98.39_{\pm0.37}$ \\
      & UCEC & $94.98$ & $96.89_{\pm1.85}$ & $97.27_{\pm1.58}$ & $97.59_{\pm0.51}$ \\
\midrule
PLCO  & BRCA & $96.42$ & $97.82_{\pm0.88}$ & $97.97_{\pm0.85}$ & $98.13_{\pm0.31}$ \\
      & COAD & $96.31$ & $98.25_{\pm0.89}$ & $98.41_{\pm0.89}$ & $98.70_{\pm0.20}$ \\
      & IHC  & $99.71$ & $99.90_{\pm0.06}$ & $99.93_{\pm0.02}$ & $99.94_{\pm0.01}$ \\
\midrule
TCGA  & BRCA & $90.05$ & $92.05_{\pm3.08}$ & $92.41_{\pm2.62}$ & $92.75_{\pm1.14}$ \\
      & COAD & $91.69$ & $95.47_{\pm2.81}$ & $95.99_{\pm2.46}$ & $96.71_{\pm0.70}$ \\
      & GBM  & $98.68$ & $99.48_{\pm0.19}$ & $99.59_{\pm0.13}$ & $99.66_{\pm0.03}$ \\
      & PAAD & $88.78$ & $92.23_{\pm3.61}$ & $92.64_{\pm3.15}$ & $93.35_{\pm1.23}$ \\
      & SKCM & $94.79$ & $97.08_{\pm0.94}$ & $97.29_{\pm1.06}$ & $97.59_{\pm0.26}$ \\
\bottomrule
\end{tabular}}
\vspace{-3mm}
\end{table}
{\setlength{\textfloatsep}{6pt plus 2pt minus 2pt}
\begin{table}[t]
\centering
\scriptsize
\setlength{\tabcolsep}{3pt}
\renewcommand{\arraystretch}{0.95}


\caption{\footnotesize\textbf{Ablation study for ZS-informed one- vs multi-shot with slide-level FM.} ID:EBRAINS $RB_0$. Values AUROC.}

\label{tab:ablation-one-vs-multishot-auroc-ebrains}
\begin{tabular}{lcccc}
\toprule
\multirow{2}{*}{\textbf{OOD Dataset}} & \multicolumn{4}{c}{\textbf{AUROC}} \\
\cmidrule(lr){2-5}
 & \textbf{1-shot} & \textbf{3-shot} & \textbf{5-shot} & \textbf{10-shot} \\
\midrule
RB$_1$ & $97.01_{\pm 0.92}$ & $97.55_{\pm0.42}$ & $97.51_{\pm0.40}$ & \textbf{$97.66_{\pm0.24}$} \\
RB$_2$ & $96.96_{\pm 1.59}$ & $97.94_{\pm0.43}$ & $97.83_{\pm0.40}$ & $97.89_{\pm 0.29}$ \\
RB$_3$ & $92.55_{\pm 1.29}$ & $93.73_{\pm 0.51}$ & $93.71_{\pm0.46}$ & $93.83_{\pm 0.29}$ \\
\bottomrule
\end{tabular}
\end{table}

\vspace{-2mm}
{\setlength{\textfloatsep}{6pt plus 2pt minus 2pt}
\begin{table}[t!]
\centering
\caption{\footnotesize\textbf{Ablation study for number of patches in the top-$j$ pooling}. ID: $RB_0$. Each cell shows AUROC for \emph{ZIO}.}
\label{tab:topj-ablation}
\resizebox{\columnwidth}{!}{%
\setlength{\tabcolsep}{3pt}
\scriptsize
\begin{tabular}{lccccc}
\toprule
\textbf{OOD} & \textbf{Top-1} & \textbf{Top-5} & \textbf{Top-10} & \textbf{Top-50} & \textbf{Top-100} \\
\midrule
RB$_1$ & $84.29_{\pm2.84}$ & $85.89_{\pm1.69}$ & $\mathbf{86.33_{\pm1.70}}$ & $85.94_{\pm1.55}$ & $84.93_{\pm1.46}$ \\
RB$_2$ & $84.46_{\pm4.61}$ & $88.13_{\pm3.08}$ & $\mathbf{88.70_{\pm2.80}}$ & $87.10_{\pm2.37}$ & $84.97_{\pm2.22}$ \\
RB$_3$ & $72.98_{\pm4.76}$ & $74.77_{\pm2.87}$ & $\mathbf{74.93_{\pm2.65}}$ & $73.11_{\pm2.51}$ & $71.23_{\pm2.25}$ \\
\bottomrule
\end{tabular}}
\end{table}



\vspace{2mm}
\noindent\textbf{Multimodal fusion strategies.} We compare the  prototype shrinkage with two other approaches for fusion of ZS and OS prototypes: \emph{(i) score\_avg} which computes average of MaxLogit scores from each prototype and \emph{(ii) score\_max} which selects the maximum of MaxLogit scores. On the challenging rare-brain cancer types with EBRAINS-RB2 as OOD set, ZIO outperforms both fusion methods: with AUROC $\mathbf{88.70}_{\pm 2.80}$ vs.\ $82.76_{\pm 2.51}$ (\emph{score\_avg}) and $76.00_{\pm 6.22}$ (\emph{score\_max}) for the patch-level model. This indicates that prototype-based fusion provides benefits beyond simple score combination. 

\section{Summary}
We introduced ZIO, a training-free multimodal OOD detector for pathology WSIs. The proposed framework can be used with any vision-language FMs in pathology, and is compatible with both slide- and patch-level formulations. We showed that the multimodal ZIO model, which integrates complementary information from text and visual prototypes, outperforms the unimodal ZS and OS baselines. ZIO achieved superior performance over 40 SOTA OOD methods across different types of domain shifts, and also showed robust generalization to external cohorts. Thanks to the training-free formulation, ZIO can be easily applied as a safeguard for various conditions, including rare diseases or stain-compatibility testing. Our study also revealed that certain types of domain shifts, such as nuanced differences between some rare cancer types, are more challenging and will require more attention in future studies. These results also underscore the importance of model testing on various, real-world OOD conditions. To support reproducibility and enable future advances, all model implementation and dataset splits are publicly released at \url{https://github.com/muskahya/ZIO}. In summary, this study represents a step towards safer deployment of AI in medicine.


\bibliographystyle{IEEEtran}
\bibliography{main}

\end{document}